\documentclass[10pt,twocolumn,letterpaper]{article}

\usepackage{iccv}
\usepackage{times}
\usepackage{epsfig}
\usepackage{graphicx}
\usepackage{amsmath}
\usepackage{amssymb}
\usepackage{booktabs}
\usepackage{tabularx}
\usepackage{multirow}
\newcolumntype{C}[1]{>{\centering\arraybackslash}p{#1}}

\def\ie{\emph{i.e.}}
\def\eg{\emph{e.g.}}
\def\etal{{\em et al.}}
\def\etc{{\em etc.}}

\newif\ifshowcomments
\showcommentstrue
\ifshowcomments
\newcommand{\TODO}[1]{{\color{red}{[TODO: #1]}}}
\newcommand{\revised}[1]{{\color[rgb]{0.2,0.7,0.2}{#1}}}

\newcommand{\phil}[1]{{\color[rgb]{0.9,0.1,0.1}{#1}}}

\else
\newcommand{\TODO}[1]{}
\newcommand{\revised}[1]{}
\newcommand{\lzhu}[1]{}
\newcommand{\phil}[1]{}
\fi

\usepackage[breaklinks=true,bookmarks=false]{hyperref}

\iccvfinalcopy 

\def\iccvPaperID{3646} 

\ificcvfinal\fi

\begin{document}

\title{Towards Accurate Alignment in Real-time 3D Hand-Mesh Reconstruction}


\author{Xiao Tang \ \ \ \ Tianyu Wang  \ \ \ \ Chi-Wing Fu\\
The Chinese University of Hong Kong\\
{\tt\small \{xtang,wangty,cwfu\}@cse.cuhk.edu.hk}
}


\maketitle
\ificcvfinal\thispagestyle{empty}\fi

\begin{abstract}
3D hand-mesh reconstruction from RGB images facilitates many applications, including augmented reality (AR).
However, this requires not only real-time speed and accurate hand pose and shape but also plausible mesh-image alignment.
While existing works already achieve promising results, meeting all three requirements is very challenging.
This paper presents a novel pipeline by decoupling the hand-mesh reconstruction task into three stages: a joint stage to predict hand joints and segmentation; a mesh stage to predict a rough hand mesh; and a refine stage to fine-tune it with an offset mesh for mesh-image alignment.
With careful design in the network structure and in the loss functions, we can promote high-quality finger-level mesh-image alignment and drive the models together to deliver real-time predictions.
Extensive quantitative and qualitative results on benchmark datasets demonstrate that the quality of our results outperforms the state-of-the-art methods on hand-mesh/pose precision and hand-image alignment.
In the end, we also showcase several real-time AR scenarios.
%
%
\end{abstract}

\section{Introduction}


\begin{figure}[!t]
	\center
	\includegraphics[width=8.25cm]{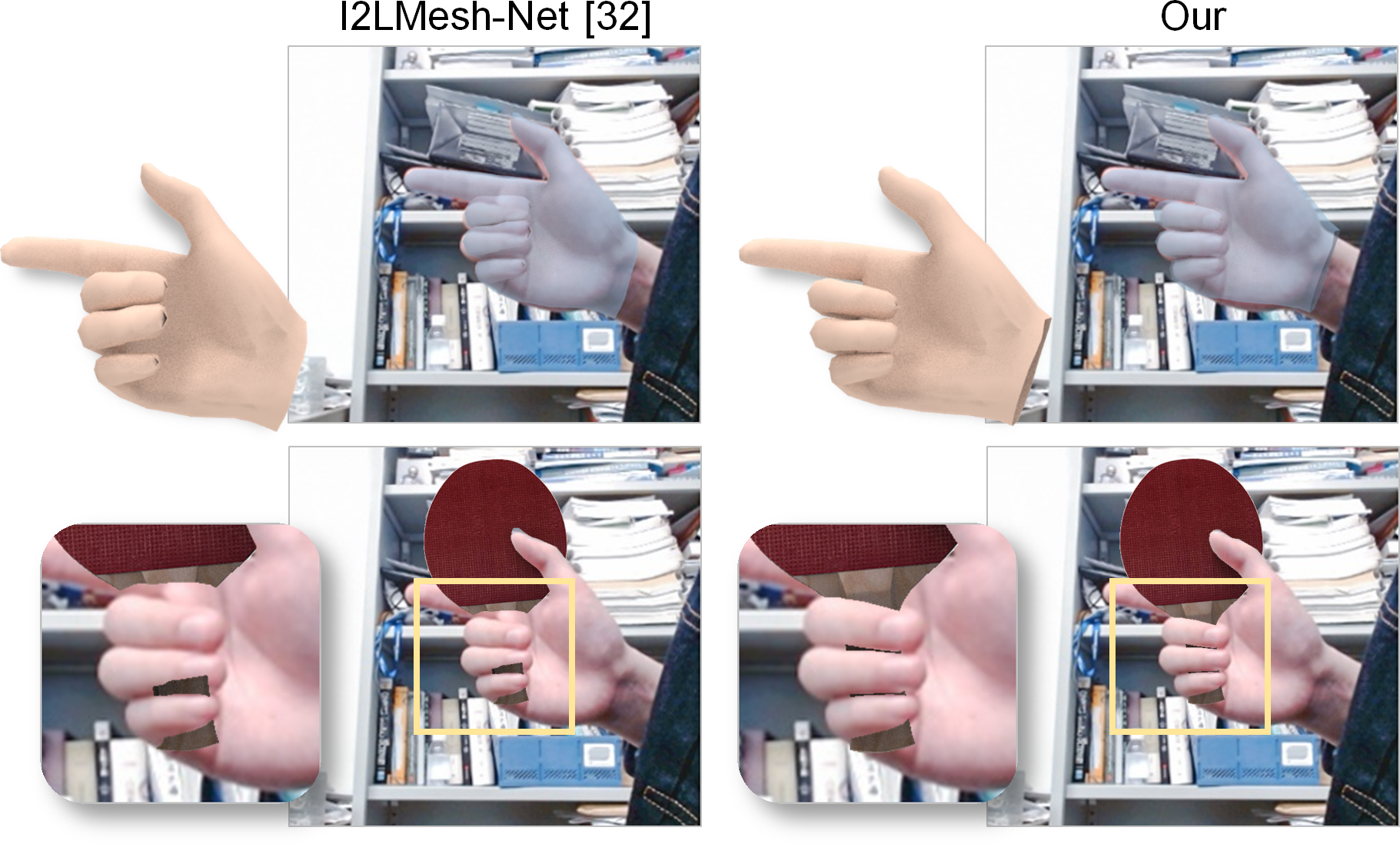}
	\caption{
	Comparing 3D hand meshes (top row) reconstructed by I2L-MeshNet~\cite{moon2020I2L} (ECCV 2020) and by our method.
	Taking these meshes to support AR interaction (bottom row), our predicted hand mesh delivers {\em better finger-level alignment\/} and {\em more natural hand occlusion\/} with the virtual ping pong racket.}
	\label{fig:teaser}
	\vspace*{-3mm}
\end{figure}

3D hand-mesh reconstruction from a single monocular view is a long-standing task in computer vision that has great potentials for supporting and enhancing many applications,~\eg, human-computer interactions, augmented reality (AR),~\etc~
By recognizing the 3D shape and pose of the user's hand in the AR view, we not only can augment the appearance of the hand and attach virtual objects onto the hand but can also enable the user to directly use his/her hand to grab and manipulate virtual 3D objects in the AR view.
These are exciting but very challenging applications in AR that require the support of computer vision methods.

To put a hand-mesh reconstruction method into practice for direct hand interactions in AR, there are three requirements to meet.
(i) The reconstruction should run in real time to give interactive feedback.
(ii) The overall pose and shape of the reconstructed hand should match the user's real hand in AR.
(iii) Beyond that, the reconstructed hand should plausibly align with the user's real hand in the image space to improve the perceptual realism in the AR interactions.

%
%

At present, research works on hand-mesh reconstruction can roughly be divided into two categories based on the type of the input image,~\ie, RGB or depth.
Despite the categorization, recent works are mostly deep-learning-based,~\eg, a typical approach is to use a deep neural network to predict the 2D/3D hand joint coordinates for guiding the regression of hand-mesh vertices~\cite{moon2020I2L} or parameters in a parametric hand model~\cite{zhou2020monocular},~\eg, MANO~\cite{romero2017embodied}.
Other works encode the image features to latent features and then use the Graph-CNN~\cite{ge20193d} or spiral filters~\cite{kulon2020weakly} to directly reconstruct the hand mesh.
Concerning the goal of supporting hand interactions in AR, existing works on hand-mesh reconstruction can mainly cater requirements (i) and (ii) but not requirement (iii), since they do not effectively utilize the 2D cues,~\eg, hand boundary, in their framework, while aiming for speed and gesture matching.
A misaligned hand mesh may lead to obvious artifacts in the AR interactions; see the left column in Figure~\ref{fig:teaser}.
While some recent works try to explore a better image-space alignment via differentiable rendering~\cite{baek2019pushing,zhang2019end} for meeting requirement (iii), they tend to struggle with requirement (i) due to the time-consuming iterative refinement process for the hand alignment.

The goal of this work is to simultaneously meet requirements (i)-(iii) for supporting real-time AR applications.
Our key idea is to decouple the hand-mesh reconstruction process into three stages,
especially to avoid the iterative refinement for hand alignment:
(i) The {\em joint stage\/} encodes the input image and uses a multi-task architecture to predict hand segmentation and joints. During the testing, we only need to compute and pass image and joint features to the next stage.
%
(ii) The {\em mesh stage\/} encodes the incoming features to quickly predict a rough 3D hand mesh with its lightweight architecture for speed.
(iii) The {\em refine stage\/} extracts local features via the local feature projection unit and global features via the global feature broadcast unit for each vertex of the mesh and then utilizes a small Graph-CNN to predict an offset mesh to quickly align the rough mesh with the user's hand in the AR image space.
In this decoupled design, offsets are essentially residuals; they are small vectors that the network can readily learn to regress based on the features gathered in the earlier stages.
Besides, we utilize differentiable rendering to promote finger-level alignment.


We conduct extensive experiments to demonstrate that our proposed method boosts the hand-mesh reconstruction accuracy, comparing with the state-of-the-art.  Also, the refine stage can efficiently produce a good image-mesh alignment for supporting AR applications.  In the end, we further showcase several AR scenarios to demonstrate the potential of our method to support real-time AR interactions.

\section{Related Work}


\noindent
{\bf 3D hand-pose estimation\/} aims to regress the 3D location of hand joints from a monocular RGB/depth image.
Recent works are mainly deep-learning-based~\cite{zimmermann2017learning,cai2018weakly,moon2018v2v,panteleris2018using,yang2019aligning,yang2019disentangling,li2019point,cai2019exploiting,chen2019so,wan2019self,baek2020weakly,fang2020jgr,huang2020awr,zhao2020knowledge,spurr2018cross,mueller2018ganerated,hasson2020leveraging}, and some~\cite{hasson2019learning,tekin2019h+,doosti2020hope} further predicted the hand-object pose.
%
%
%
Among them, Li~\etal~\cite{li2019point} designed a binary selector that divides joints into groups and learns features independently for each group to avoid negative transfer.
%
Cai~\etal~\cite{cai20203d} trained a network with both synthetic data and weakly-labeled real RGB images.
Very recently, Transformer~\cite{vaswani2017attention} was adopted to further boost the hand-pose estimation precision~\cite{he2020epipolar,huang2020hand}.



\vspace*{1mm}
\noindent
{\bf Depth-based 3D hand-mesh reconstruction\/} aims to create the hand mesh from a depth image.
Earlier works~\cite{tan2016fits,taylor2014user,khamis2015learning} often fitted a deformable hand mesh to the depth image with an iterative optimization.
Recently, deep learning further helped to improve the performance.
Malik~\etal~\cite{malik2018deephps} adopted a CNN to regress the parameters of a linear-blend skinning model.
Wan~\etal~\cite{wan2020dual} mapped features from a depth image to a mesh grid then sampled the mesh grid to recover the hand mesh.
Malik~\etal~\cite{malik2020handvoxnet} voxelized the depth image and applied 3D convolutions to reconstruct the hand mesh.
On the other hand, Mueller~\etal~\cite{mueller2019real} tracked bimanual hand interactions with a depth camera in real time.


\noindent
{\bf RGB-based 3D hand-mesh reconstruction\/} aims to reconstruct hand meshes using a commodity RGB camera~\cite{yang2020seqhand,moon2020deephandmesh,fan2020adaptive,qian2020html,han2020megatrack,wang2020rgb2hands,choutas2020monocular,shen2020phong,yang2020bihand,chen2021cameraspace}.
One common approach was to train a deep neural network with a parametric statistical model like MANO~\cite{romero2017embodied}, which parameterizes a given triangular hand mesh with pose and shape parameters.
%
Boukhayma~\etal~\cite{boukhayma20193d} regressed the MANO and view parameters to project the predicted MANO model to image space for supervision.
Zhang~\etal~\cite{zhang2019end} and Baek~\etal~\cite{baek2019pushing} adopted a differentiable rendering approach to supervise the training for silhouette alignment.
Zhou~\etal~\cite{zhou2020monocular} first predicted the 3D coordinates of hand joints, then took the joint predictions as prior to produce the hand mesh via an inverse kinematics network.
Moon~\etal~\cite{moon2020I2L} introduced an image-to-lixel network to enhance the reconstruction accuracy.

To explicitly encode mesh structure in a deep neural network, Graph-CNN was widely adopted for hand-mesh reconstruction. Yet, Graph-CNN is designed for aggregating adjacent features based on the mesh topology, so it is less efficient for long-range features.
To overcome this drawback, Ge~\etal~\cite{ge20193d} and Choi~\etal~\cite{choi2020pose2mesh} proposed to regress the hand-mesh vertices using Graph-CNN in a coarse-to-fine manner.
More recently, Kulon~\etal~\cite{kulon2020weakly} applied spiral filters for neighbourhood selection.
Spurr~\etal~\cite{spurr2020weakly} proposed biomechanical constraints with weak supervision to effectively leverage additional 2D annotated images. 



Existing works on hand-mesh reconstruction mostly focus on improving the precision of gesture prediction without efficiently utilizing 2D cues from the image.
So these works are subpar for the hand-image alignment, which is, in fact, crucial for AR applications.
In this work, we propose to decouple the hand-mesh estimation reconstruction, such that each stage can focus on a specific task for effective network learning and lightweight architecture.
In this way, our approach can efficiently produce good-quality 3D hand meshes that align with the user's hand in the AR view, while delivering real-time performance.



\begin{figure*}[t]
	\center
	\includegraphics[width=0.995\textwidth]{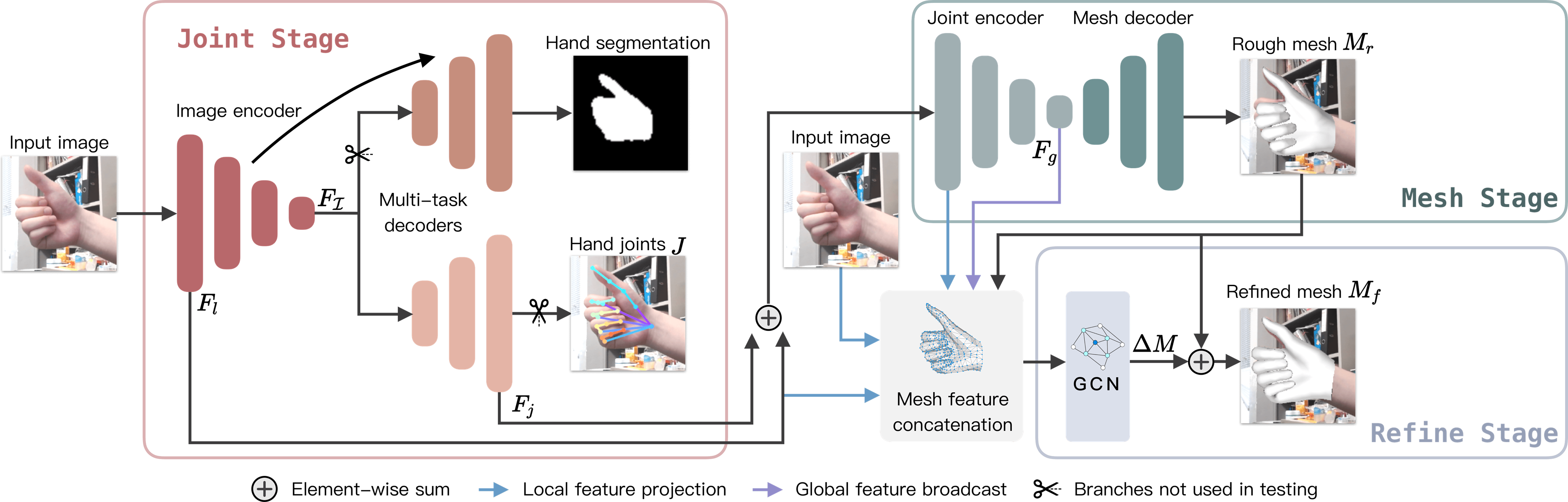}
	\vspace*{1.5mm}
	\caption{Our proposed framework. From the input image with the user's hand, the joint stage first extracts feature map $F_{\mathcal{I}}$ and sends it to two decoders, one for hand segmentation and the other for generating joint feature $F_j$ and predicting the hand-joint locations $J$.
	Subsequently, the mesh stage fuses $F_j$ and the shallow image feature $F_l$ from the joint stage for predicting rough hand mesh $M_r$.
	Lastly, we aggregate image feature $F_l$ from the joint stage with the shallow layers of the joint encoder and global feature $F_g$ from the mesh stage, and passes the combined feature and the rough mesh to the graph convolutional layers (GCN) in the refine stage to regress offset mesh $\Delta M$ and produce the final hand mesh $M_f = M_r + \Delta M$.
	The scissor icon means the corresponding branch can be cut off during the testing. 
}
	\label{fig:network}
\end{figure*}
\section{Method}

Figure~\ref{fig:network} shows our three-stage framework for reconstructing the hand mesh from an input RGB photo:
%
\begin{itemize}
\vspace*{-1.25mm}
\item[(i)]
\textbf{The joint stage} encodes the input image and predicts a hand segmentation mask and hand joints $J$.
\vspace*{-1.25mm}
\item[(ii)]
\textbf{The mesh stage} encodes features from the previous stage, including joint feature $F_j$ from the joint decoder and shallow image feature $F_l$ from the image encoder, and then predicts rough hand mesh $M_r$.
\vspace*{-1.25mm}
\item[(iii)]
\textbf{The refine stage} aggregates the projected local and global features from the previous stages with the vertices in the rough mesh, then adopts a Graph-CNN (GCN) to regress offset mesh $\Delta M$ for producing the final hand mesh $M_f = M_r + \Delta M$.
%
\end{itemize}
%
Each stage in our framework has a clear goal, so they can {\em better focus on learning the associated features\/}, e.g., joint stage for the overall hand shape and joints, mesh stage for the rough hand mesh, and refine stage for learning to regress the {\em offset vectors to align the rough mesh with the user's hand in the image space\/}.
%
%
Also, the network model in each stage can be kept {\em small and compact for achieving real-time performance\/}.
%
%
Next, we present the three stages in detail.


\subsection{Joint Stage}
\label{subsec:joint_stage}
Given the input image, the joint stage first encodes it using a feature extractor and then feeds the encoded feature $F_{\mathcal{I}}$ into two branches: {\em one to predict the hand joints\/} and {\em the other to predict the hand segmentation\/}.
For the hand-joint branch, we use a feature decoder to generate the joint feature map $F_j$ and then regress the hand-joint locations,~\ie, 3D coordinates of the 21 joints of a hand (denoted as $J$), from $F_j$ and $F_{\mathcal{I}}$ via multiple dilated convolutional layers and soft-argmax~\cite{chapelle2010gradient}, following a similar strategy as~\cite{moon2020I2L}:
%
%
%
%
\begin{equation}
\left\{ \begin{array}{rcl}
J_x & = & \mathrm{soft}\textrm{-}\hspace*{-0.6mm}\arg\max(\mathrm{Conv_{1D}}({avg^x(F_j)})),\\
J_y & = & \mathrm{soft}\textrm{-}\hspace*{-0.6mm}\arg\max(\mathrm{Conv_{1D}}({avg^y(F_j)})),\\
J_z & = & \mathrm{Conv_{1D}}(\Phi({avg^{x,y}(F_{\mathcal{I}})})),
\end{array} \right.
\end{equation}
where subscripts $x$ and $y$ in $J$ denote the image space and $z$ denotes depth;
$\mathrm{Conv_{1D}}$ denotes 1D convolution;
$\Phi$ denotes a block, which consists of a fully connection layer, 1D batch normalization, and reshape function that rearranges the feature vector from $\mathbb{R}^C$ to $\mathbb{R}^{c'\times d}$;
$C$ is the channel size of feature vector;
$d$ is the size of $avg^x(F_l)$;
and $c' = C / d$.

Considering that we need a plausible alignment between the reconstructed hand mesh and the user's hand in the image space, we thus harass shallow features from the image encoder to better capture the fine details of the hand, especially at the boundaries.
Hence, we introduce another prediction head in the joint stage for the hand segmentation (see Figure~\ref{fig:network}), in which we use a U-Net~\cite{ronneberger2015u} to process the encoded feature map and predict the hand segmentation.
%



\subsection{Mesh Stage}

In the mesh stage, we first fuse joint feature $F_j$ and shallow image feature $F_l$ from the joint stage through a joint encoder; see again Figure~\ref{fig:network}.
Note that we choose $F_j$ instead of $J$ as input, since the predicted joints $J$ may not be highly accurate, so it may misguide the reconstruction.
Also, combining shallow feature $F_l$ helps preserve the image features.
%
%
%
The fused feature is then sent to an encoder unit to produce global mesh feature $F_g$.
After that, we regress the 3D coordinates of the vertices in hand mesh $M_r$ from $F_g$ through multiple dilated convolutional layers and soft-argmax.

Since the rough mesh is restored from a low-resolution (1/64$\times$1/64 of the original) global feature, in which most local features have lost due to the dilated kernels, so the result is mostly smooth and undersampled at boundaries.
So, the rough mesh may not well align with the user's hand in the input image.
Yet, this stage can quickly generate a rough hand mesh in 3D that captures the overall hand shape, while the refine stage can fine-tune the mesh by regressing per-vertex offset vectors that are small and easy to learn.



\subsection{Refine Stage}

The bottom right of Figure~\ref{fig:network} illustrates how we residually refine the rough mesh $M_r$ predicted by the previous mesh stage. Inspired by~\cite{kirillov2020pointrend, kolotouros2019convolutional,wang2018pixel2mesh}, we design the {\em local feature projection unit\/} and the {\em global feature broadcast unit\/} to further extract local and global features from the input image and previous encoder layers, respectively.

\begin{figure}[t]
	\center
	\includegraphics[width=6.0cm]{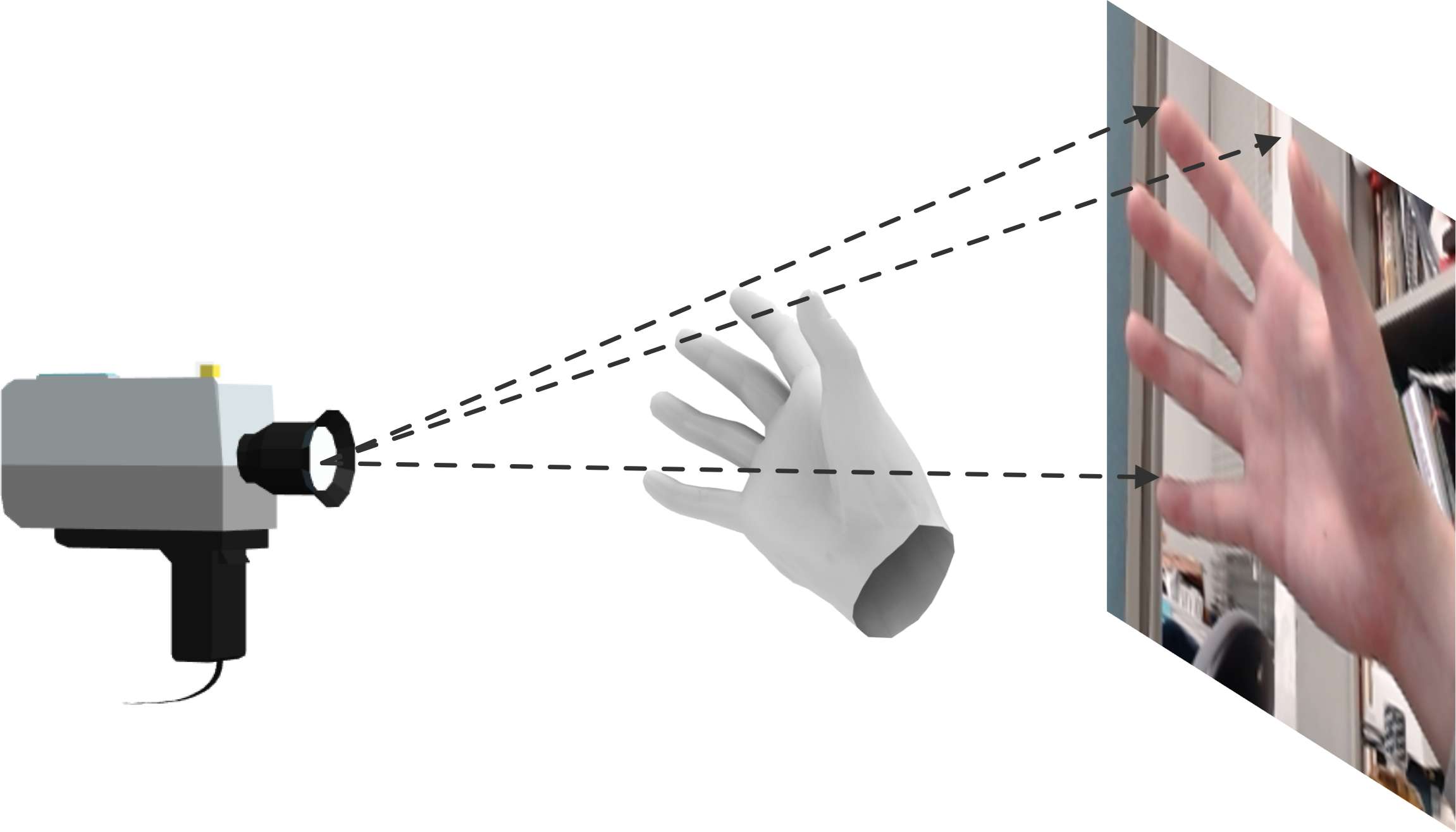}
	\caption{
	Through a 3D-to-2D projection from the 3D hand mesh to the 2D image space, we can extract high-resolution image features in the image feature space with a bilinear interpolation and collect image features for each vertex in the rough hand mesh.}
	\label{fig:local}
\end{figure}

\vspace*{1mm}
\noindent
{\bf Local feature projection unit.} \
To produce a fine hand mesh that aligns with the real hand in input image, we are inspired by~\cite{kirillov2020pointrend}~to design the local feature projection 
unit to extract a single local feature vector per vertex in the rough mesh based on the previous shallow layers and input image.

Similiar to~\cite{wang2018pixel2mesh}, we first project each mesh vertex to the 2D image space of the input image (equivalently, the spatial domain of feature maps) via a 3D-to-2D projection, as illustrated in Figure~\ref{fig:local}.
Then, we perform a bilinear interpolation around each projected vertex on the feature maps to extract associated feature vectors. 
Here, the local feature projection unit collects features from the input image, the first layer of the image encoder, and also the first layer of the joint encoder; see the light blue arrows in Figure~\ref{fig:network}.

\vspace*{1mm}
\noindent
{\bf Global feature broadcast unit.} \
The local features help us to address the fine details but they are insufficient.
It is because they do not provide (global) information about the overall mesh structure. 
Concerning this, we introduce the global feature broadcast unit that broadcasts the mesh's global feature to every mesh vertex.  Here, inspired by~\cite{kolotouros2019convolutional}, we take deep feature $F_g$ from the joint encoder in the mesh stage, apply a global average pooling over it to obtain a single 1-D vector, and reduce its channel dimension by 1/4 using a fully connected layer.
After that, we attach this global feature vector to every vertex of the rough mesh; see the light purple arrow from $F_g$ in Figure~\ref{fig:network}.

\begin{figure}[t]
	\center
	\includegraphics[width=8.15cm]{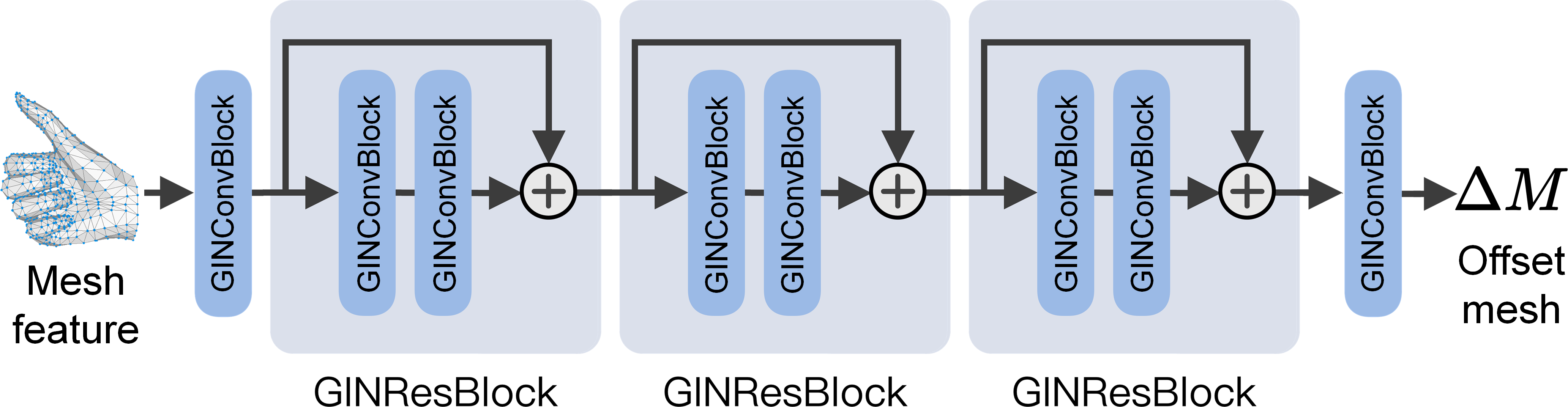}
	\caption{An illustration of our Graph-CNN network, whose input is the concatenated mesh feature (see Figure~\ref{fig:network}).
	This network starts with a GINConvBlock, followed by three GINResBlocks and another GINConvBlock, to produce the offset mesh $\Delta M$.}
	\label{fig:gcn}
\end{figure}

\vspace*{1mm}
\noindent
{\bf Graph-CNN.} \
After collecting the local features from the local feature projection unit and global features from the global feature broadcast unit, we concatenate all the features at each vertex
in the mesh and then send the concatenated mesh feature
to our Graph-CNN.
From a high-level perspective, the Graph-CNN aims to estimate one 3D offset vector for each mesh vertex based on its input 3D coordinates, as well as the collected local and global features, for aligning the rough mesh to the hand in the image.
This processing is done by propagating features over the hand mesh topology. 
For the graph convolution layers, we adopt the formulation of Graph Isomorphism Network (GIN) convolution from~\cite{xu2018how}, which is defined as:
\begin{equation}
\begin{aligned}
{x}'_i = \textrm{MLPs}(x_i + \sum_{j\in\mathcal{N}(i)}{x_j}),
\end{aligned}
\end{equation}
where $x_i$ is the feature of the $i$-th node in the graph (equivalently the $i$-th vertex in the rough mesh);
$x'_i$ is the updated feature of the node;
$\mathcal{N}(i)$ is an index set of the neighboring nodes of the $i$-th node;
and MLPs denotes a series of multi-layer perceptrons.
The framework of the Graph-CNN is illustrated in Figure~\ref{fig:gcn}.
The network starts with a GINConvBlock, which contains a GIN convolution layer with ReLU and 1D batch normalization.
After that, the network employs three GINResBlocks~\cite{kolotouros2019convolutional}, each with two GINConvBlocks and an identity connection, followed by a GINConvBlock, to generate the offset mesh $\Delta M$.

\subsection{Training \& Inference}
We denote a dataset as $\{I^i, S^i, J_{gt}^i, M_{gt}^i\}^{N}_{i=1}$, where $N$ is the total number of samples in the dataset;
$I^i$ is the $i$-th input image;
$S^i$ is the associated binary mask of the hand; and
$J_{gt}^i$ and $M_{gt}^i$ are the associated annotated 3D joint and mesh coordinates of the hand, respectively.
Note that the 3D joint coordinates can also be calculated from the mesh using a pre-defined regression matrix $G$,~\eg, $J_{gt}^i = GM_{gt}^i$.

We adopt an L1 loss to formulate both the mesh loss $\mathcal{L}_{\textrm{mesh}}$ and joint loss $\mathcal{L}_{\textrm{joint}}$ to supervise the predictions of mesh and joint in our three-stage framework:
%
\begin{equation}
\begin{aligned}
\mathcal{L}_{\textrm{mesh}} \ = \
\mid\mid M_{gt}^i - M_r^i \mid\mid + 
\mid\mid M_{gt}^i - M_f^i \mid\mid,
\end{aligned}
\end{equation}
%
\begin{equation}
\begin{aligned}
\mathcal{L}_{\textrm{joint}} \ = \
\mid\mid J_{gt}^i - J^i \mid\mid + 
\mid\mid J_{gt}^i - GM_r^i \mid\mid \
\\
+
\mid\mid J_{gt}^i - GM_f^i \mid\mid,
\end{aligned}
\end{equation}
Besides, inspired by~\cite{wang2018pixel2mesh}, we adopt the normal loss $\mathcal{L}_{\textrm{norm}}$ for preserving the surface normals and the edge-length loss $\mathcal{L}_{\textrm{edge}}$ to penalize flying vertices:
%
\begin{equation}
\begin{aligned}
\mathcal{L}_{\textrm{norm}} = \sum_{f \in M_{gt}^i}\sum_{e \in f}{\mid\mid \langle \vec{e_r}, n_{gt}^{f} \rangle \mid\mid}\\ +
\sum_{f \in M_{gt}^i}\sum_{e \in f}{\mid\mid \langle \vec{e_f}, n_{gt}^{f} \rangle \mid\mid}
\end{aligned}
\end{equation}
\begin{equation}
\begin{aligned}
\text{and} \ \
\mathcal{L}_{\textrm{edge}} = \sum_{f \in M_{gt}^i}\sum_{e \in f}{\mid\mid \lvert \vec{e_f}\rvert - \lvert \vec{e_{gt}}\rvert \mid\mid}\\ +
\sum_{f \in M_{gt}^i}\sum_{e \in f}{\mid\mid \lvert \vec{e_r}\rvert - \lvert \vec{e_{gt}}\rvert \mid\mid},
\end{aligned}
\end{equation}
where $f$ denotes a triangle face;
$e$ denotes an edge of the triangle;
$\vec{e_{gt}}$, $\vec{e_r}$, and $\vec{e_f}$ denote the edge vector on $f$ that comes from $M_{gt}^i$, $M_r^i$, and $M_f^i$, respectively; and
$n_{gt}^{f}$ denotes the surface normal of $f$ based on $M_{gt}^i$.

Next, we adopt the standard cross-entropy loss to supervise the hand segmentation:
\begin{equation}
\begin{aligned}
\mathcal{L}_{\textrm{sil}} = -\sum^{H \times W}_{j} y^j\log{p^j},
\end{aligned}
\end{equation}
where $H \times W$ denotes the size of $S^i$;
$y^j$ denotes pixel $j$ of $S^i$;
and $p^j$ denotes the prediction result of pixel $j$.

Lastly, we employ a differentiable renderer to render the predicted final hand mesh $M_f$ in the image space to supervise the alignment through the render loss $\mathcal{L}_{\textrm{render}}$. Different from~\cite{zhang2019end,baek2019pushing} that use a binary silhouette mask in the supervision, we paint fingers of the ground-truth hand mesh with different colors and supervise the prediction by color matching to promote finger-level recognition.
As the colors in the painted image indicate the occlusion relationship between the fingers and palm, ``render'' can further improve the joint prediction accuracy as well.
Formally, we have
\begin{equation}
\begin{aligned}
\mathcal{L}_{\textrm{render}} = \frac{1}{H_\mathcal{R} \times W_\mathcal{R}}\sum^{H_\mathcal{R} \times W_\mathcal{R}}_{j}\mid\mid\mathcal{R}(M_f^i)^j - \mathcal{R}(M_{gt}^i)^j \mid\mid^2,
\end{aligned}
\end{equation}
where $\mathcal{R}$ denotes the differentiable renderer; $H_{\mathcal{R}} \times W_{\mathcal{R}}$ denotes the output resolution of $\mathcal{R}$;
and
$\mathcal{R}(M_f^i)^j$ and $\mathcal{R}(M_{gt}^i)^j$ denotes the color of pixel $j$ in $\mathcal{R}(M_f^i)$ and in $\mathcal{R}(M_{gt}^i)$, respectively.

Our overall loss is
$
\mathcal{L} = \mathcal{L}_{\textrm{mesh}} + \lambda_j\mathcal{L}_{\textrm{joint}} +
\lambda_n\mathcal{L}_{\textrm{normal}} +
\lambda_e\mathcal{L}_{\textrm{edge}} +
\lambda_s\mathcal{L}_{\textrm{sil}} + 
\lambda_r\mathcal{L}_{\textrm{render}}
$,
where we empirically set $\lambda_j=\lambda_n=\lambda_e=1$, $\lambda_s=10$, and $\lambda_r=0.1$.

During the inference time, we can cut off the hand segmentation branch and the hand joints prediction in the joint stage to save computing time, without lowering the performance; see the scissor icons in Figure~\ref{fig:network}.
\section{Experiment \& Results}

\subsection{Experimental Settings}

\paragraph*{Datasets.} 
(i) FreiHAND~\cite{zimmermann2019freihand} contains 32,560 real training samples
and 3,960 real test samples, 
all annotated with the MANO model.
We use FreiHAND for both training and testing.
Note that we randomly selected 2,000 samples from the training set for alignment evaluation and employed the rest for training. 
(ii) ObMan~\cite{hasson2019learning} has 141,550 synthetic training samples rendered with the MANO model. We also train our network on ObMan to further boost its performance.
(iii) EgoDexter~\cite{mueller2017real} has 1,485 real samples captured in an egocentric view; we use it only for testing.

\begin{figure*}[t]
	\center
\includegraphics[width=0.99\linewidth]{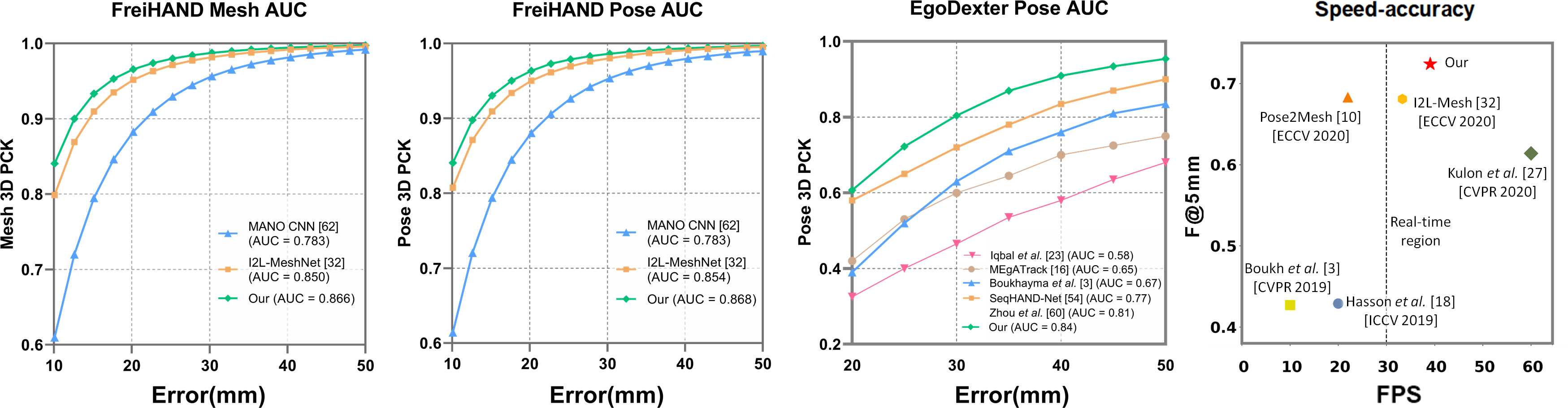}
	\caption{The mesh/pose AUC comparison with the state-of-the-art methods.
	From left to right, the first and second plots present the mesh and pose AUC results on the FreiHAND dataset, respectively, whereas the third plot presents the pose AUC result on the EgoDexter dataset.
	The fourth plot presents the speed-accuracy plot of 3D hand-mesh reconstruction on the FreiHAND test-set, in which we obtain the inference time by testing all methods on Nvidia RTX2080Ti-GPU. 
	Our method achieves the best performance, while being real-time.}
	\label{fig:auc}
	\vspace{-2mm}
\end{figure*}

\vspace{-2mm}
\paragraph*{Evaluation metrics.}
For quantitative evaluation, we follow the evaluation metrics in the FreiHAND online competition.
(i) Mesh/pose error (ME/PE) measures the Euclidean distances between the predicted and ground-truth mesh vertices and joint coordinates.
(ii) Mesh/Pose AUC reports the area under the curve of the percentage of correct key points (PCK) curve in different error threshold ranges.
(iii) F-scores measures the harmonic mean of the recall and precision between the predicted and ground-truth vertices; here, we follow existing works to use F@5mm and F@15mm.

To evaluate image-space alignment, we compare the projected hand-mesh silhouette with the ground truth on two common segmentation metrics:
(i) Mean Intersection over Union (mIoU), which measures the overlap regions and
(ii) Hausdoff distance (HD), which is the maximum distance of a set to the nearest point in another set, so it is more sensitive at boundaries.
We report $95\%$ HD in the evaluation.

\vspace{-2mm}
\paragraph*{Implementation details.}
We used ResNet-50~\cite{he2016deep} pretrained on ImageNet~\cite{russakovsky2015imagenet} as the joint/mesh encoders.
Our network was trained using the Adam optimizer on two Nvidia RTX Titan GPUs with a batch size of $32$ per GPU for $25$ epochs and an initial learning rate of $1e^{-4}$ (decay rate of $0.1$ per $10$ epochs).
We resized the input image to $256$$\times$$256$ and augmented the data with random scaling, rotation, and color jittering.
Our network runs at 39.1 frames per sec. on Nvidia RTX 2080 Ti, after we cut off the hand segmentation and joint prediction branches in joint stage.
{\em Our code is released at \small{\url{https://wbstx.github.io/handar}}.\/}

\begin{figure}[t]
	\center
	\includegraphics[width=8.05cm]{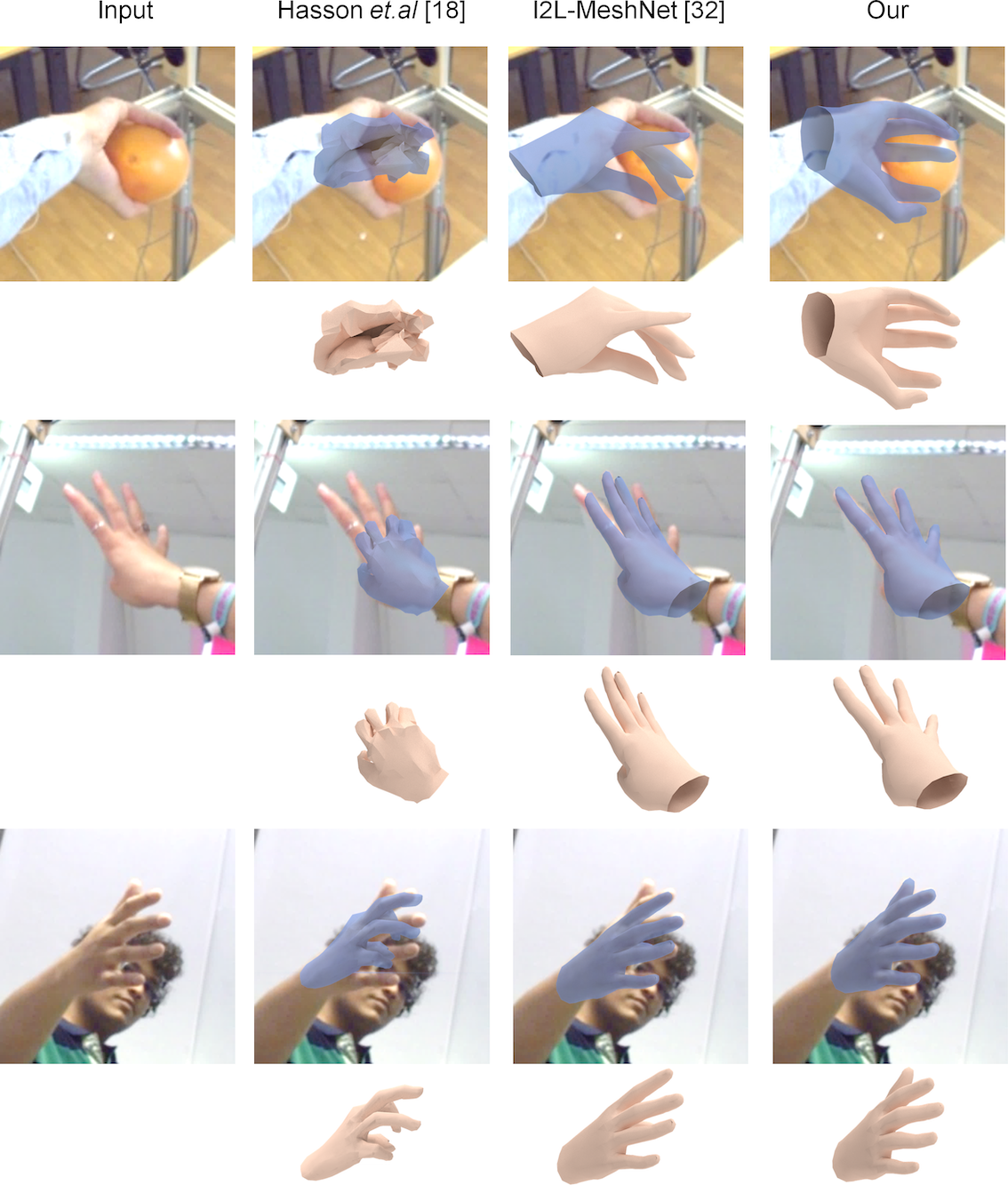}
	\caption{Visual comparisons between our method and state-of-the-arts.
	The input images (left column) are from the FreiHAND dataset~\cite{zimmermann2019freihand}.
	Comparing the other three columns, we can see that our method predicts better hand meshes that match the gestures in input images with high-quality finger-level alignment.}
	\label{fig:datasetComparison}
	\vspace{-2mm}
\end{figure}


\begin{figure*}[t]
	\center
\includegraphics[width=0.99\linewidth]{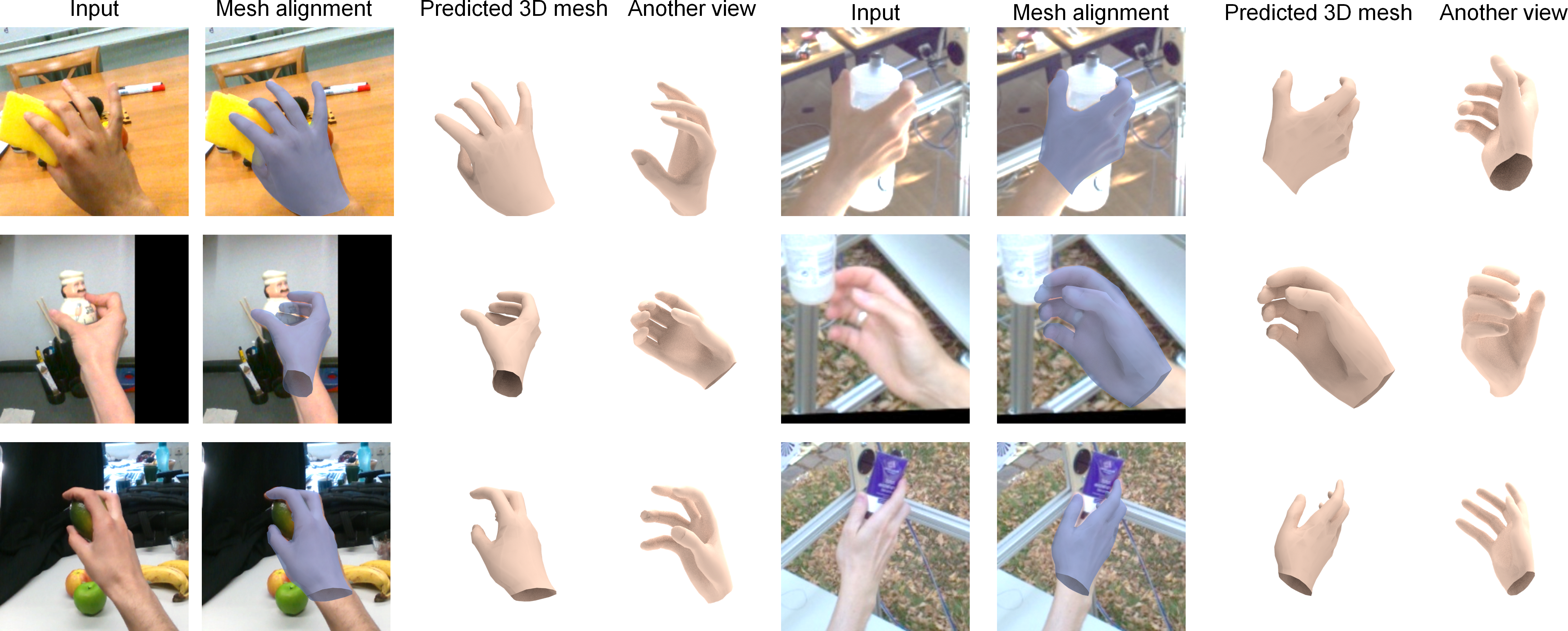}
	\caption{More qualitative results on test images in the EgoDexter dataset (left) and in the FreiHAND dataset (right).}
	\label{fig:gallery}
\end{figure*}

\begin{figure}[t]
       \center
       \includegraphics[width=0.8\linewidth]{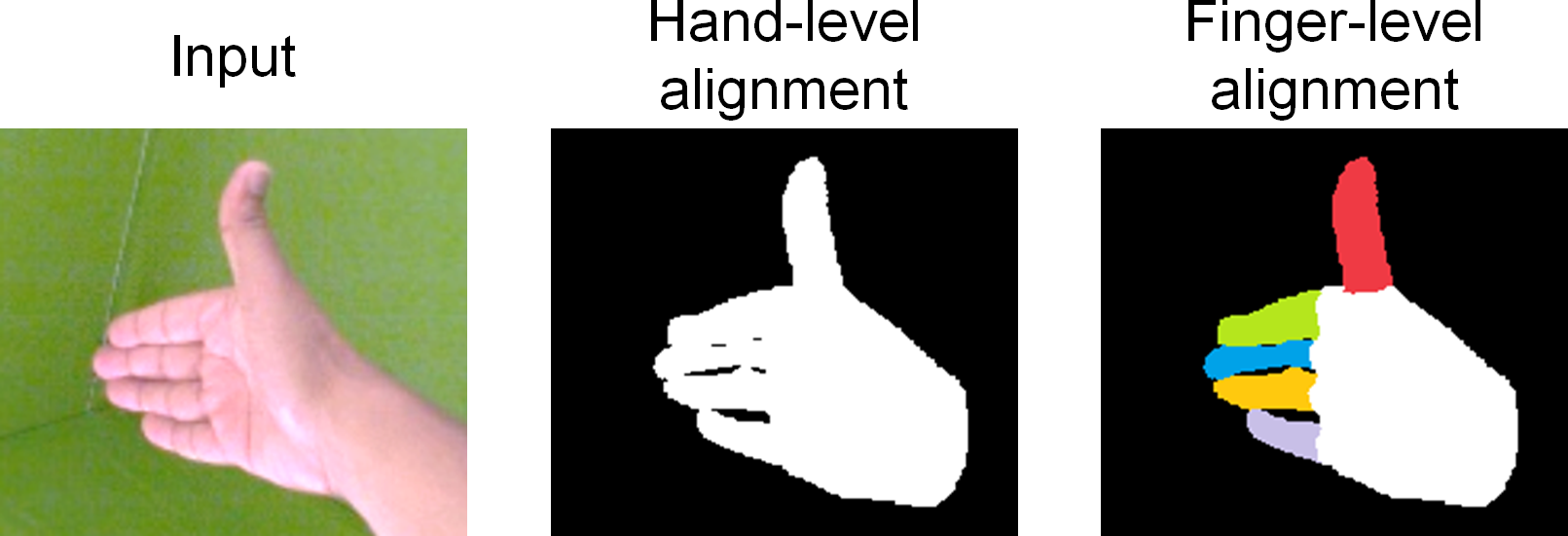}
       \caption{The alignment evaluation is conducted on two levels. For hand-level alignment, we evaluate only the silhouette of the whole hand between the prediction and ground truth. For finger-level alignment, we evaluate individual fingers and palm. Finger-level alignment is more demanding than hand-level alignment.}
       \label{fig:alignment}
\end{figure}

\begin{table}[t]
	\centering
	\caption{Quantitative comparisons between our method and state-of-the-arts on the FreiHAND dataset.
	$\downarrow$ means the lower the better; 
	$\uparrow$ means the higher the better; and 
	* means the method requires 2D joint annotations as input.
	The unit of ME and PE is 10mm.}
	\vspace{2mm}
	\resizebox{\linewidth}{!}{
		\begin{tabular}{@{\hspace{1mm}}c@{\hspace{1mm}}c@{\hspace{1mm}}||
				@{\hspace{2mm}}c@{\hspace{2mm}}c@{\hspace{2mm}}c@{\hspace{2mm}}c@{\hspace{1mm}}c
			} \toprule[1pt]
			Methods & venue & ME $\downarrow$ & PE $\downarrow$ & F@5mm $\uparrow$ & F@15mm $\uparrow$ & FPS\\ \hline
			Mean shape~\cite{zimmermann2019freihand}
			& ICCV2019
			&1.64 & 1.71  & 0.376  & 0.873   & -\\
			MANO fit~\cite{zimmermann2019freihand}
			& ICCV2019
			&1.37 & 1.37  & 0.439  & 0.892   & -\\
			Hasson~\etal~\cite{hasson2019learning}
			& ICCV2019
			&1.33 & 1.33  & 0.429  & 0.907   & 20\\
			Boukh.~\etal~\cite{boukhayma20193d}
			& CVPR2019
			&1.32 & 3.50  & 0.427  & 0.894   & 10\\
			ExPose~\cite{choutas2020monocular}
			& ECCV2020
			&1.22 & 1.18  & 0.484  & 0.918   & -\\
			MANO CNN~\cite{zimmermann2019freihand}
			& ICCV2019
			&1.09 & 1.10  & 0.516  & 0.934   & -\\
			Kulon~\etal~\cite{kulon2020weakly}
			& CVPR2020
			&0.86 & 0.84  & 0.614  & 0.966  & 60\\
			I2L-MeshNet~\cite{moon2020I2L}
			& ECCV2020
			&0.76 & 0.74  & 0.681  & 0.973  & 33.3\\
			Pose2Mesh*~\cite{choi2020pose2mesh}
			& ECCV2020
			&0.76 & 0.74  & 0.683  & 0.973  & 22\\ \hline
			Our w/o ObMan~\cite{hasson2019learning}&
			& 0.71  & 0.71  & 0.706 & 0.977 & 39.1  \\ 
			Our w/ ObMan~\cite{hasson2019learning}&
			&\textbf{0.67} & \textbf{0.67} & \textbf{0.724} & \textbf{0.981} & 39.1
			\\ \bottomrule[1pt]
	\end{tabular}}
	\label{tab:quanComparison}
\end{table}

\subsection{Comparison with State-of-the-art Methods}
We follow existing works~\cite{zimmermann2019freihand,hasson2019learning,boukhayma20193d,choutas2020monocular,kulon2020weakly,moon2020I2L,choi2020pose2mesh} to quantitatively compare our method with state-of-the-art methods on the FreiHAND test set.
Since the ground truths of the test set are not accessible, the evaluation is conducted by submitting our test results to the online server.
Table~\ref{tab:quanComparison} and the left two plots of Figure~\ref{fig:auc} report the results, showing that our method outperforms others for all the aforementioned metrics.
Further, Figure~\ref{fig:datasetComparison} shows visual comparisons on three test images in the dataset. For each input image, the first row shows the hand-image alignment by each method and the second row shows the predicted hand meshes.
Please note that the code of Kulon~\etal~\cite{kulon2020weakly} has not been released and Pose2Mesh~\cite{choi2020pose2mesh} requires 2D joints annotations as input, we omit them in the comparison.
%
%
Comparing the results, we can see that other methods may fail to predict correct gestures or not able to well align the hand meshes in the image space.
Our method is able to predict more accurate hand poses and shapes, and well aligns the hand meshes with the hands in images.
More comparisons can be found in the supplemental material.

Next, we evaluate on an unseen dataset (not used in training), which is EgoDexter~\cite{mueller2017real}, to compare the generality of our method with several existing pose estimation methods~\cite{iqbal2018hand, boukhayma20193d,zhou2020monocular,han2020megatrack,yang2020seqhand}.
Following~\cite{zhou2020monocular}, we use the centroids of the finger tips as roots to align the prediction and the ground truth. The third plot in Figure~\ref{fig:auc} shows the AUC result. Note that, since Zhou~\etal~\cite{zhou2020monocular} did not provide their pose PCK curve on EgoDexter, we only report their AUC.
On the other hand,~\cite{moon2020I2L} did not test their performance on the EgoDexter dataset, so we do not report it in the figure.
%
From the plot, we can see that our method achieves the highest pose AUC value, comparing with all the other methods, demonstrating its generality and potential for practical usage.
Please check Figure~\ref{fig:gallery} for more qualitative results from EgoDexter and FreiHAND. Also, we report the speed-accuracy comparison in the fourth plot shown in Figure~\ref{fig:auc}. As we can see from this plot and also from Table~\ref{tab:quanComparison}, our method beats
all
recent methods on mesh prediction quality and also achieves real-time performance. 

\subsection{Comparison of Alignment}
To further evaluate the alignment quality, we render the predicted hand mesh to produce a silhouette mask and compare it with the ground-truth silhouette using the aforementioned metrics.
Here, we compare our method with I2L-MeshNet~\cite{moon2020I2L}, which achieves state-of-the-art performance on the FreiHAND dataset without requiring 2D joint annotations.
Since there are no ground-truth masks for the FreiHAND test set, we randomly
select 2,000 images from the training set for validation and evaluate the hand-image alignment with resolution $224 \times 224$ on two levels:
(i) hand-level alignment, in which we only evaluate the binary silhouette of the hand; and
(ii) finger-level alignment, in which we evaluate the alignment for each finger and the palm.
Note that finger-level alignment is more demanding than hand-level alignment; see Figure~\ref{fig:alignment} for illustrations.

Table~\ref{tab:alignment} shows the quantitative comparison result, from which we can see that our method outperforms I2LMesh-Net~\cite{moon2020I2L} with a much higher mIoU and smaller HD values, which reveals a much better overall image-mesh alignment and also the accuracy at the boundaries.

\begin{table}[t]
	\centering
	\caption{Quantitative alignment comparisons between our method and I2L-MeshNet~\cite{moon2020I2L}. $\downarrow$ means the lower the better and $\uparrow$ means the higher the better.
	Note that we report the average HD for the fingers and palm in the evaluation of finger-level alignment.}
	\vspace{2mm}
	\resizebox{7.0cm}{!}{
		\begin{tabular}{@{\hspace{1mm}}c@{\hspace{1mm}}||
				@{\hspace{2mm}}c@{\hspace{2mm}}c@{\hspace{2mm}}||c@{\hspace{2mm}}c
			} \toprule[1pt]
			\multirow{2}*{Methods} 
			& \multicolumn{2}{c||}{Hand-level}
			& \multicolumn{2}{c}{Finger-level} \\
			\cline{2-5} 
			& mIoU $\uparrow$ & HD $\downarrow$ & mIoU $\uparrow$ & HD $\downarrow$ \\ \hline
			I2L-MeshNet~\cite{moon2020I2L}
			&92.08 & 6.17 & 71.07 & 8.42 \\
			Our w/o ObMan~\cite{hasson2019learning} & 92.86 & 4.91 & 77.08 & 7.06\\
			Our w/ ObMan~\cite{hasson2019learning} &\textbf{92.95} & \textbf{4.70} & \textbf{77.33} & \textbf{6.82}
			\\ \bottomrule[1pt]
	\end{tabular}}
	\label{tab:alignment}
\end{table}

\subsection{Ablation Study}

\begin{figure}[]
	\center
	\includegraphics[width=0.98\linewidth]{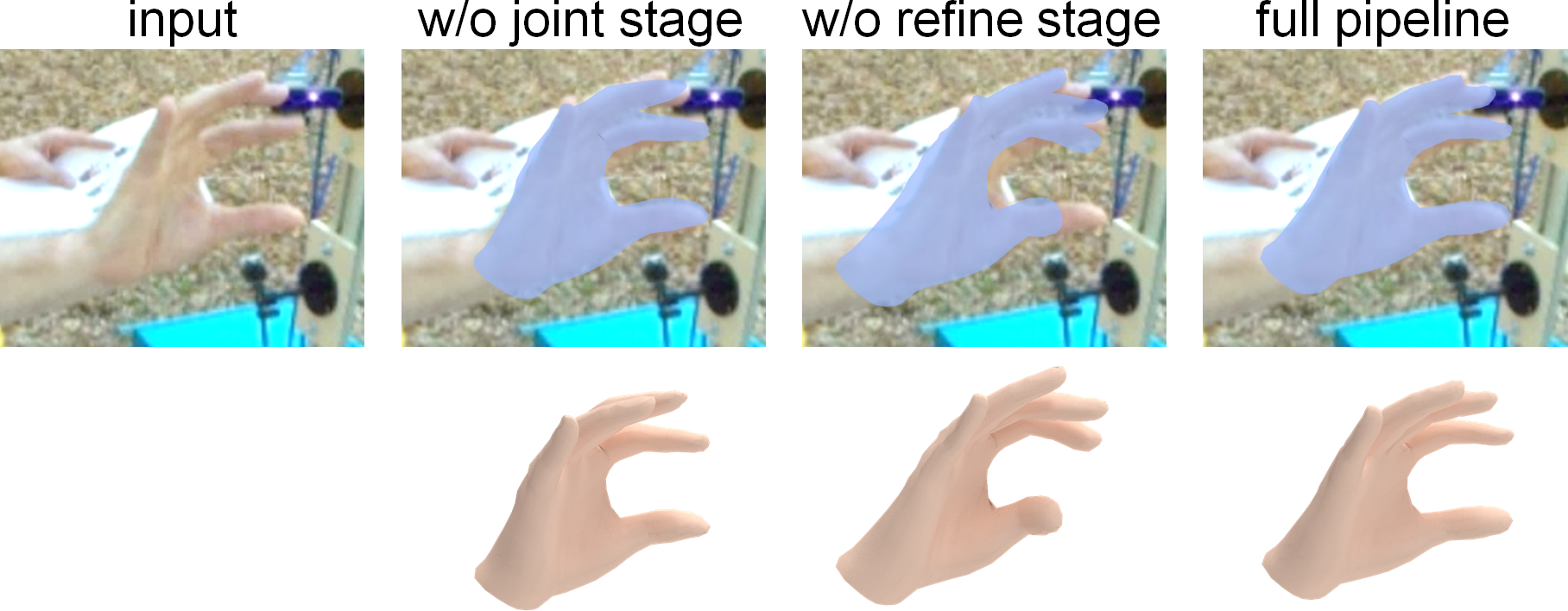}
	\caption{Ablation study on our decoupled design.}
	\label{fig:ablation}
\end{figure}

\paragraph*{Effects of decoupled design.} To evaluate our decoupled design, we conducted an ablation study by simplifying our pipeline for the following two cases: (i) remove the joint stage and directly predict hand segmentation in the mesh stage; and (ii) remove the refine stage and treat $M_r$ as the final hand mesh. The ablation results are reported in the top two rows of Table~\ref{tab:ablation}. We can observe obvious performance drop when removing any one of the two stages, showing the effectiveness of each stage and their contributions to the whole framework for better performance.
Note that we conducted three runs and reported the average for each ablation case.
Please see Figure~\ref{fig:ablation} for visual comparison.

\begin{table}[t]
\caption{Comparing the performance of our full pipeline (bottom-most) with various ablation cases. For each case in the experiment, we conducted three runs and reported the average.}
\centering
	\vspace{2mm}
	\resizebox{8.25cm}{!}{
		\begin{tabular}{@{\hspace{1mm}}c@{\hspace{1mm}}||
			@{\hspace{1mm}}c@{\hspace{1mm}}c@{\hspace{1mm}}ccc
			} \toprule[1pt]
			Models & Mesh AUC
			 $\uparrow$ & Pose AUC $\uparrow$ & mIoU$\uparrow$ & HD$\downarrow$ \\ \hline
			w/o joint stage & 0.858{$\scriptstyle \pm$0} & 0.860{$\scriptstyle \pm$0} & 92.42{$\scriptstyle \pm$0.10} & 5.12{$\scriptstyle \pm$0.08}\\
			w/o refine stage & 0.860{$\scriptstyle \pm$0} & 0.861{$\scriptstyle \pm$0} & 92.05{$\scriptstyle \pm$0.04} & 5.93{$\scriptstyle \pm$0.07}\\
			w/o local & 0.860${\scriptstyle \pm 5e^{-4}}$ & 0.862${\scriptstyle \pm 5e^{-4}}$  & 92.55{$\scriptstyle \pm$0.05} & 5.34{$\scriptstyle \pm$0.04}\\
			w/o global & 0.862${\scriptstyle \pm 5e^{-4}}$ & 0.864${\scriptstyle \pm 5e^{-4}}$  & 92.63{$\scriptstyle \pm$0.04} & 5.01{$\scriptstyle \pm$0.07}\\
			w/o offset & 0.782${\scriptstyle \pm 2e^{-3}}$ & 0.858${\scriptstyle \pm 3e^{-3}}$  & 88.01{$\scriptstyle \pm$0.10} & 7.97{$\scriptstyle \pm$0.11}\\
			w/o $\mathcal{L}_{\textrm{sil}}$ & 0.862${\scriptstyle \pm 5e^{-4}}$  & 0.864${\scriptstyle \pm 5e^{-4}}$   & 92.50{$\scriptstyle \pm$0.08} & 5.19{$\scriptstyle \pm$0.08}\\
			w/o $\mathcal{L}_{\textrm{render}}$ & 0.863{$\scriptstyle \pm$0} & 0.865{$\scriptstyle \pm$0} & 92.80{$\scriptstyle \pm$0.02} & 4.97{$\scriptstyle \pm$0.04}\\ \hline
			Full & \textbf{0.866}${\scriptstyle \pm 5e^{-4}}$ & \textbf{0.868}${\scriptstyle \pm 5e^{-4}}$ & \textbf{92.95}{$\scriptstyle \pm$0.04} & \textbf{4.70}{$\scriptstyle \pm$0.02}\\
			\bottomrule[1pt]
	\end{tabular}}
	\label{tab:ablation}
	\vspace*{-3mm}
\end{table}

\begin{figure*}[t]
	\center
	\includegraphics[width=0.92\linewidth]{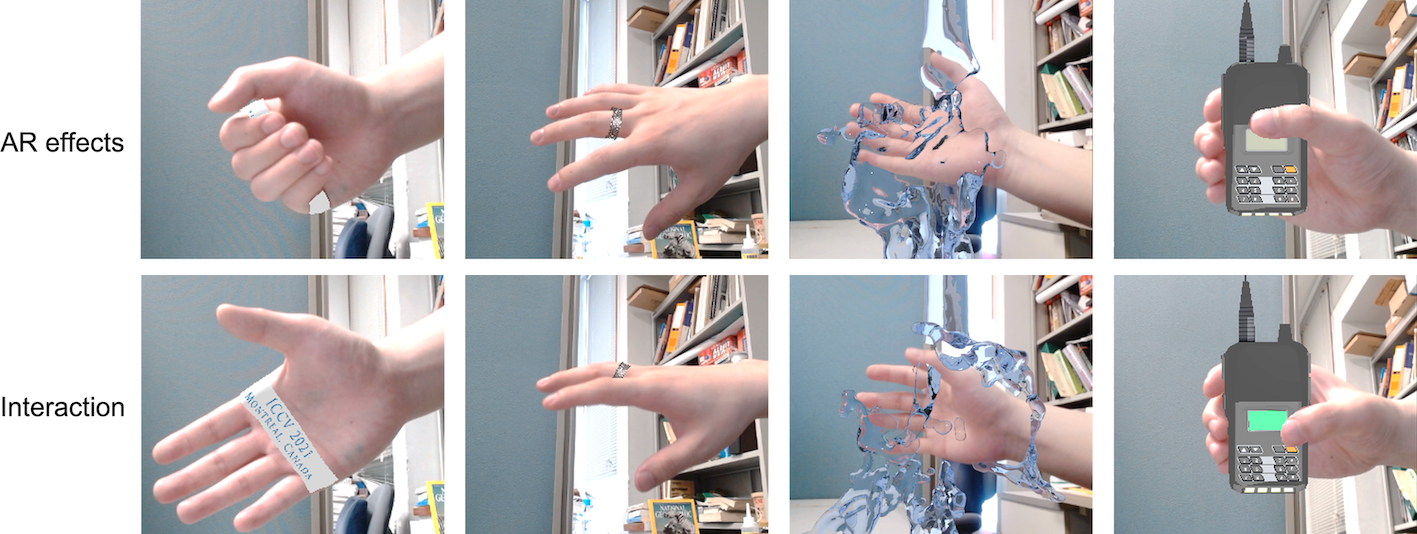}
	\caption{We showcase four example AR interaction scenarios to demonstrate our method's applicability (from left to right):
	(i) a virtual paper band around the user's hand,
	(ii) a virtual try-on with a ring (note the plausible hand occlusions with the virtual objects);
	(iii) we may splash virtual water on the user's hand and run a physical simulation, and
	(iv) the user can directly grab and manipulate a virtual walkie-talkie, and press a button on it.
	{\em Please watch full videos of these scenarios in~\footnotesize{\url{https://wbstx.github.io/handar}}.}}
	\label{fig:application}
	\vspace*{-1mm}
\end{figure*}

\paragraph*{Effects of refinement unit \& loss terms.} 
Further, we evaluate the contributions of various units in the refine stage and in the losses for the following cases:
(i) remove the local projection unit;
(ii) remove the global broadcast unit;
(iii) directly regress final mesh vertices by the Graph-CNN instead of offset vectors;
(iv) remove the segmentation branch and $\mathcal{L}_{\textrm{sil}}$; and
(v) remove $\mathcal{L}_{\textrm{render}}$.
Table~\ref{tab:ablation} reports the ablation results, in which we also conducted three runs and reported the average for each ablation case.
Clearly, our full pipeline performs the best for all metrics, and removing any component reduces the overall performance, showing that each component contributes to improve the final result.
Particularly, note that our method has a large performance drop when using the Graph-CNN to directly regress the mesh vertices instead of offset vectors. The reason behind is that our Graph-CNN is designed to regress small values, which are relatively easy to learn for this small Graph-CNN.

\subsection{Applications}
Our method can support direct hand interactions with 3D virtual objects in AR.
Figures~\ref{fig:teaser} and~\ref{fig:application} show various interaction scenarios. With the reconstructed hand mesh that plausibly aligns with the user's hand in the input image, we can
(i) resolve the occlusion between the virtual objects and the user's hand (see the first two examples in the figure);
(ii) interact with a physical simulation, e.g., water, in the 3D AR space (see the third example); and
(iii) directly grab and manipulate a virtual object, say by pressing on a button in the virtual walkie-talkie (see the last example in the figure).
The associated full videos that were lively captured can be found in the supplemental material.

\section{Conclusion \& Future Work}
This paper presents a new framework for 3D hand-mesh reconstruction by decoupling the task into three stages: the joint stage predicts the 3D coordinates of hand joints and the hand segmentation mask; the mesh stage estimates a rough 3D hand mesh; and the refine stage collects local and global features from previous layers and learns to regress per-vertex offset vectors to help align the rough mesh to the hand image with finger-level alignment. Experimental results demonstrate that our method outperforms the state-of-the-art methods for both the hand mesh/pose precision and mesh-image alignment.  Also, our method is fast and can run in real-time on commodity graphics hardware.

In the future, we plan to explore methods to produce high-resolution (fine-grained) hand-mesh predictions
that better match the smooth boundaries of the real hand.
%
Also, we would like to explore the possibility of designing an adaptive refine stage that focuses more on aligning boundary vertices to improve the overall efficiency, while further cutting off the computation. Lastly, we plan also to explore the deployment of our method on mobile devices.
\vspace*{-3mm}
\paragraph*{Acknowledgments.} 
We thank anonymous reviewers for the valuable comments. This work is supported by the Research Grants
Council of the Hong Kong Special Administrative Region (Project No. CUHK 14206320).

{\small
\bibliographystyle{ieee_fullname}
\bibliography{egbib}
}

\end{document}